\documentclass[twocolumn,english]{article}
\usepackage[T1]{fontenc}
\usepackage[latin9]{inputenc}
\usepackage{color}
\usepackage{array}
\usepackage{refstyle}
\usepackage{booktabs}
\usepackage{url}
\usepackage{multirow}
\usepackage{amsmath}
\usepackage{amssymb}
\usepackage{graphicx}
\PassOptionsToPackage{normalem}{ulem}
\usepackage{ulem}

\makeatletter

\AtBeginDocument{\providecommand\Tabref[1]{\ref{Tab:#1}}}
\AtBeginDocument{\providecommand\Eqref[1]{\ref{Eq:#1}}}
\AtBeginDocument{\providecommand\Figref[1]{\ref{Fig:#1}}}
\AtBeginDocument{\providecommand\Secref[1]{\ref{Sec:#1}}}
\AtBeginDocument{\providecommand\parref[1]{\ref{par:#1}}}
\providecommand{\tabularnewline}{\\}
\RS@ifundefined{subsecref}
  {\newref{subsec}{name = \RSsectxt}}
  {}
\RS@ifundefined{thmref}
  {\def\RSthmtxt{theorem~}\newref{thm}{name = \RSthmtxt}}
  {}
\RS@ifundefined{lemref}
  {\def\RSlemtxt{lemma~}\newref{lem}{name = \RSlemtxt}}
  {}

\PassOptionsToPackage{hyphens}{url}
\usepackage{cvpr}
\usepackage{booktabs, pifont}
\usepackage{tikz}
\usepackage[table]{xcolor}
\usepackage{colortbl}
\usepackage{graphicx}
\usepackage[accsupp]{axessibility}  
\newcommand{\good}{\textcolor{green!65!black}{\ding{51}}}
\newcommand{\bad}{\textcolor{red!80!black}{\ding{55}}}

\definecolor{cvprblue}{rgb}{0.21,0.49,0.74}
\usepackage[pagebackref,breaklinks,colorlinks,allcolors=cvprblue]{hyperref}

\def\paperID{7577} 
\def\confName{CVPR}
\def\confYear{2026}

\title{FrameDiT: Diffusion Transformer with Matrix Attention for\\ Efficient Video Generation}

\def\spaces{~~~~~~}
\author{
Minh Khoa Le$^{1}$\spaces{}Kien Do$^{2}$\spaces{}Duc Thanh Nguyen$^{3}$\spaces{}Truyen Tran $^{1}$\\
$^{1}$ Applied Artificial Intelligence Initiative, Deakin University, Australia\\
$^{2}$ FPT Smart Cloud, Vietnam \spaces{} $^{3}$ Deakin University, Australia \\
$^{1,3}$ \emph{\{minh.le, duc.nguyen, truyen.tran\}@deakin.edu.au} \\
$^{2}$ \emph{kiendd6@fpt.com}
}

\usepackage{varwidth}

\makeatother

\usepackage{babel}
\begin{document}
\maketitle

\global\long\def\Real{\mathbb{R}}%
\global\long\def\Expect{\mathbb{E}}%
\global\long\def\Normal{\mathcal{N}}%
\global\long\def\video{x}%
\global\long\def\latent{z}%
\global\long\def\query{q}%
\global\long\def\key{k}%
\global\long\def\mvalue{v}%
\global\long\def\Model{\text{FrameDiT}}%
\global\long\def\Modelglobal{\text{FrameDiT-G}}%
\global\long\def\Modeluniversal{\text{FrameDiT-H}}%

\begin{abstract}
High-fidelity video generation remains challenging for diffusion models
due to the difficulty of modeling complex spatio-temporal dynamics
efficiently. Recent video diffusion methods typically represent a
video as a sequence of spatio-temporal tokens which can be modeled
using Diffusion Transformers (DiTs). However, this approach faces
a trade-off between the strong but expensive Full 3D Attention and
the efficient but temporally limited Local Factorized Attention. To
resolve this trade-off, we propose Matrix Attention, a frame-level
temporal attention mechanism that processes an entire frame as a matrix
and generates query, key, and value matrices via matrix-native operations.
By attending across frames rather than tokens, Matrix Attention effectively
preserves global spatio-temporal structure and adapts to significant
motion. We build $\Modelglobal$, a DiT architecture based on Matrix
Attention, and further introduce $\Modeluniversal$, which integrates
Matrix Attention with Local Factorized Attention to capture both large
and small motion. Extensive experiments show that $\Modeluniversal$
achieves state-of-the-art results across multiple video generation
benchmarks, offering improved temporal coherence and video quality
while maintaining efficiency comparable to Local Factorized Attention.

\end{abstract}

\section{Introduction \label{sec:Introduction}}

\begin{table}[t]
\centering{}\caption{Comparison of our proposed Global and Hybrid (Global--Local) factorized
attention mechanisms (bold) with existing attention designs for DiT.
The symbols \good, \bad$\ $indicate whether each method possesses
a given property.\label{tab:intro-comparison}}
{\footnotesize{}}%
\begin{tabular}{>{\raggedright}m{2.2cm}cccc}
\toprule 
\multirow{2}{2.2cm}{{\small{}Property}} & \multirow{2}{*}{\textcolor{black}{\small{}Full 3D}} & \multicolumn{3}{c}{{\small{}Spatio-Temporal Factorized}}\tabularnewline
\cmidrule{3-5} \cmidrule{4-5} \cmidrule{5-5} 
 &  & {\small{}Local} & \textbf{\small{}Global} & \textbf{\small{}Hybrid}\tabularnewline
\midrule 
{\small{}Handling }{\small\par}

{\small{}Large Motion} & {\small{}\good} & {\small{}\bad} & {\small{}\good} & {\small{}\good}\tabularnewline
\midrule 
{\small{}Computationally Efficient} & {\small{}\bad} & {\small{}\good} & {\small{}\good} & {\small{}\good}\tabularnewline
\midrule 
{\small{}Fusing Global }{\small\par}

{\small{}and Local} & {\small{}\bad} & {\small{}\bad} & {\small{}\bad} & {\small{}\good}\tabularnewline
\bottomrule
\end{tabular}
\end{table}

\begin{figure*}[!tp]
\begin{centering}
\includegraphics[width=0.84\textwidth]{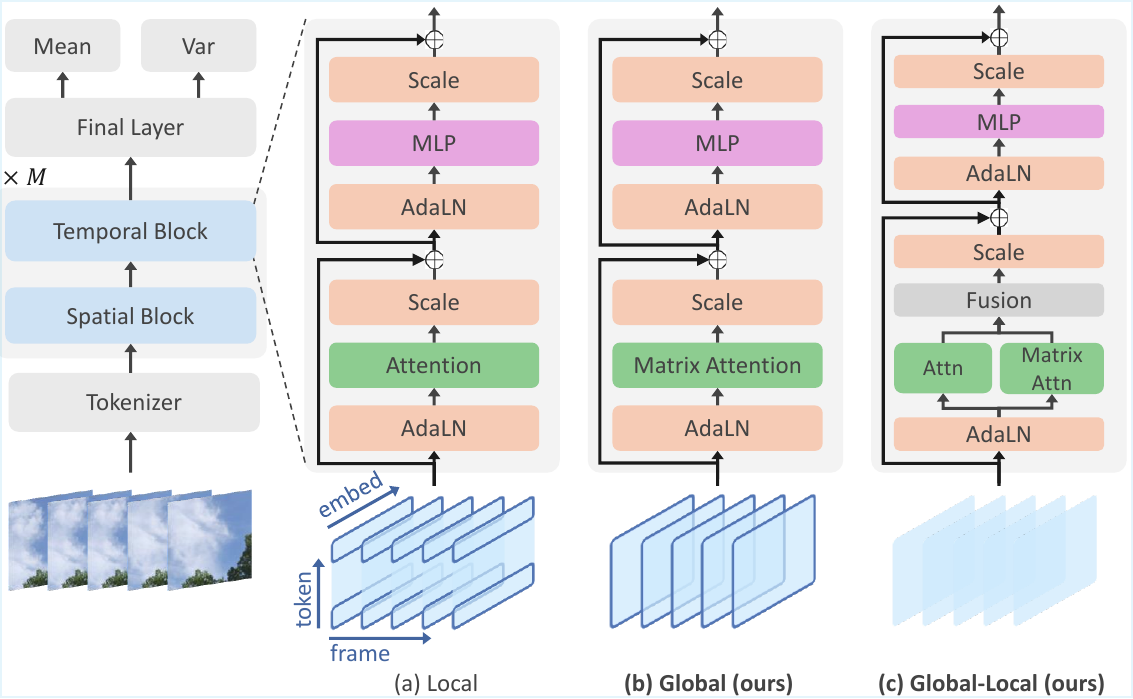}
\par\end{centering}
\caption{\textbf{\label{fig:overview}Overview of the proposed $\boldsymbol{\protect\Model}$.}
Built on the Diffusion Transformer with interleaved Spatial and Temporal
blocks. (a) Local: conventional local factorized attention; (b) Global
(ours): replaces temporal attention with Matrix Attention for frame-level
temporal attention; (c) Global--Local Hybrid (ours): combines local
and global temporal attention for unified spatio-temporal modeling.}
\end{figure*}

Generative models have witnessed remarkable progress in recent years.
Among various approaches, Diffusion Models \cite{song2019generative,ho2020denoising}
have emerged as the dominant paradigm, achieving success in synthesizing
high-fidelity images that are indistinguishable from real \cite{nichol2021glide,rombach2022high,mokady2023null,nguyen2025h,bdbm,kieu2026universal}.
Extending this success from static images to videos represents the
next major frontier, with transformative impact across content creation,
filmmaking, and world model \cite{vdm2022,rolling,chen2024diffusion,ma2025latte,le2025newton,le2025gravityvideogenerationposttraining,le2026piscesannotationfreetexttovideoposttraining}.

A video is not simply a collection of independent frames but a temporally
coherent sequence governed by complex spatio-temporal relationships.
The main challenge in video generation is therefore to model these
dependencies both effectively and efficiently. Early video diffusion
models \cite{vdm2022,blattmann2023stable,mei2023vidm} rely on 3D
U-Nets with factorized spatiotemporal layers. More recent works \cite{hong2022cogvideo,zheng2024open,yang2025cogvideox,song2025history,ma2025latte,wan2025wan,sun2025ar}
adopt the Diffusion Transformer (DiT) architecture \cite{peebles2023scalable},
which scales better with model size and support long-range temporal
modeling. Within DiT-based models, attention mechanisms generally
fall into two main categories:
\begin{enumerate}
\item \textbf{Full 3D Attention} \cite{kong2024hunyuanvideo,hacohen2024ltx,lin2024open,sun2025ar,song2025history,yang2025cogvideox,wan2025wan}:
video is treated as a sequence of $T\times N$ tokens, and joint spatial-temporal
attention is applied. While highly expressive, its computational complexity
grows quadratically as $O\left(T^{2}N^{2}\right)$, making it expensive
for high-resolution or long-duration video synthesis.
\item \textbf{(Spatially) Local Factorized Attention }\cite{hong2022cogvideo,cong2023flatten,lu2023vdt,zheng2024open,ma2025latte}:
spatial attention is first applied within each frame, followed by
temporal attention across frames for each spatial location. This reduces
complexity to $O\left(T^{2}N+TN^{2}\right)$. However, the temporal
attention only links tokens at corresponding spatial positions, making
these models struggle to capture large motions, where objects do not
remain spatially aligned across frames.
\end{enumerate}
These two designs expose a clear trade-off between expressiveness
and computational efficiency, motivating the question: \emph{Can we
design a DiT architecture that captures temporal coherence as effectively
as Full 3D Attention while remaining as efficient as factorized attention?}

To answer this question, we propose \emph{Matrix Attention}, a novel
attention mechanism for video modeling that operates at the frame
level rather than the token level. It treats each input frame as a
matrix, where rows correspond to tokens and columns correspond to
token embedding dimensions, and uses matrix-native operations to compute
query, key, and value matrices for attending over other frames. As
a result, Matrix Attention effectively captures global spatio-temporal
structure and remains robust to large motion - unlike the spatially
local temporal attention used in prior work \cite{hong2022cogvideo,cong2023flatten,lu2023vdt,zheng2024open,ma2025latte}.

We integrate Matrix Attention into DiT to create $\Model$, a new
factorized video diffusion transformer with two temporal block variants:
Global version that uses only Matrix Attention, and Global-Local Hybrid
version that additionally incorporates standard temporal attention
for fine-grained modeling. As summarized in \Tabref{intro-comparison},
our model achieves the \textquotedbl best of both worlds\textquotedbl :
it achieves the expressiveness of Full 3D Attention while maintaining
the computational efficiency of Local Factorized Attention. 

We evaluate $\Model$ across a wide range of video generation benchmarks,
including UCF-101, Sky-Timelapse, Taichi-HD (128/256), and FaceForensics.
The Global variant consistently improves temporal coherence with minimal
computational overhead, while the Hybrid variant achieves state-of-the-art
FVD and FVMD on multiple datasets. We further analyze long video generation,
model size scaling, data efficiency, and runtime/memory characteristics,
demonstrating that Matrix Attention maintains efficiency comparable
to factorized attention while improving the temporal consistency.

In summary, our contributions are:
\begin{itemize}
\item Matrix Attention - a novel frame-level temporal attention mechanism
that captures the global spatio-temporal structure in videos.
\item $\Modelglobal$ - a spatio-temporally factorized DiT architecture
built on Matrix Attention for video diffusion models, and $\Modeluniversal$
- an enhanced hybrid version that integrates local factorized attention
to jointly model global and local motion.
\item Extensive experiments demonstrating the effectiveness and efficiency
of $\Modelglobal$ and $\Modeluniversal$ compared to existing baselines.
\end{itemize}

\section{Preliminaries}

\subsection{Diffusion Models}

\begin{figure*}[!tp]
\begin{centering}
\resizebox{0.95\linewidth}{!}{ \includegraphics[width=1\textwidth]{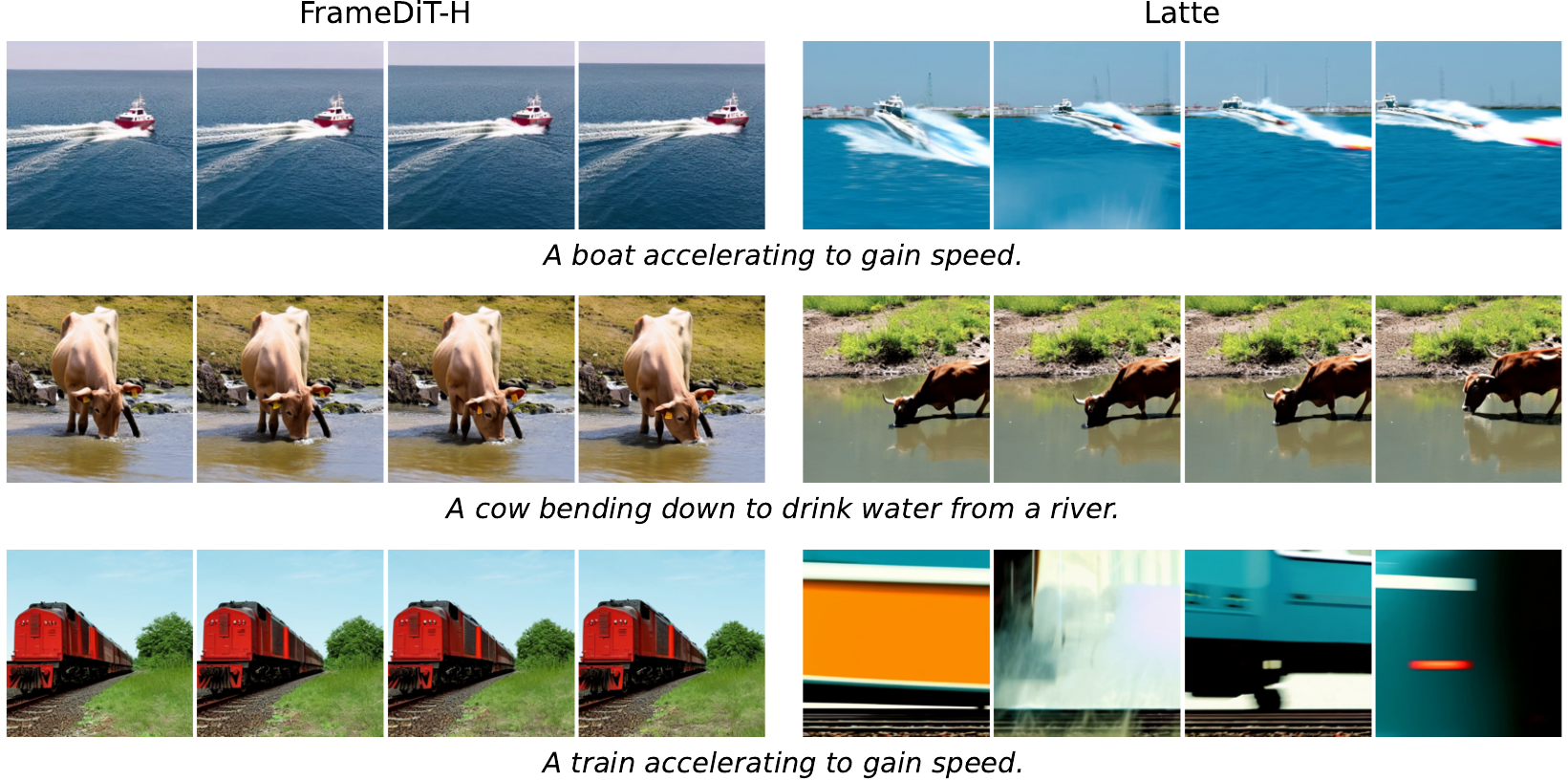}}
\par\end{centering}
\caption{\textbf{Text-to-video generation comparison between Latte and our
$\textbf{\ensuremath{\protect\Modeluniversal}}$.} We show 4 of 16
generated frames.\label{fig:t2v-qualitative}}
\end{figure*}

Diffusion models \cite{ho2020denoising} generate data by reversing
a forward diffusion process. The forward process is a Markov chain
with Gaussian transition kernels $q\left(x_{k}|x_{k-1}\right)$ that
gradually transforms a clean data sample $x\sim p\left(x\right)$
into Gaussian noise $x_{K}\sim\mathcal{N}\left(0,I\right)$. The transition
is designed so that $q\left(x_{k}|x\right)$ has the form $\mathcal{N}\left(a_{k}x,\sigma_{k}^{2}I\right)$,
allowing to sample $x_{k}$ directly from $x$ as:
\begin{equation}
x_{k}=a_{k}x+\sigma_{k}\epsilon,\ \;\epsilon\sim\mathcal{N}\left(0,I\right),\label{eq:forward}
\end{equation}
where $a_{k}$, $\sigma_{k}$ are non-negative functions of $k$ with
a decreasing signal-to-noise ratio $\frac{a_{k}^{2}}{\sigma_{k}^{2}}$.
When $a_{k}^{2}+\sigma_{k}^{2}=1$, the model is variance-preserving
\cite{song2020score}, as in DDPM \cite{ho2020denoising}, where $a_{k}=\sqrt{\bar{\alpha}_{k}}$
and $\sigma_{k}=\sqrt{1-\bar{\alpha}_{k}}$. The reverse process is
defined by parameterized Gaussian transitions $p_{\theta}\left(x_{k-1}|x_{k}\right)=\mathcal{N}\left(\mu_{\theta,k,k-1}\left(x_{k}\right),\omega_{k|k-1}^{2}\mathrm{I}\right)$,
with mean:
\begin{align}
\mu_{\theta,k,k-1}\left(x_{k}\right) & =\frac{a_{k-1}}{a_{k}}x_{k}\label{eq:reverse-mean}\\
 & +\left(\sqrt{\sigma_{k-1}^{2}-\omega_{k-1|k}^{2}}-\frac{\sigma_{k}a_{k-1}}{a_{k}}\right)\epsilon_{\theta}\left(x_{k},k\right),\nonumber 
\end{align}
and the variance $\omega_{k|k-1}^{2}:=\eta^{2}\sigma_{k-1}^{2}\left(1-\frac{\sigma_{k-1}^{2}}{\sigma_{k}^{2}}\frac{a_{k}^{2}}{a_{k-1}^{2}}\right)$
($\eta\in\left[0,1\right]$). The noise network $\epsilon_{\theta}\left(x_{k},k\right)$
is trained to recover the Gaussian noise $\epsilon$ used in the forward
process (see \Eqref{forward}) by minimizing the noise matching (NM)
loss:
\begin{equation}
\mathcal{L}_{\text{NM}}\left(\theta\right)=\Expect_{x,k,\epsilon}\left[\left\Vert \epsilon_{\theta}\left(x_{k},k\right)-\epsilon\right\Vert ^{2}\right]\label{eq:noise-matching}
\end{equation}
where $x\sim p\left(x\right)$, $k\sim\text{Uniform}\left(1,K\right)$,
and $\epsilon\sim\mathcal{N}\left(0,\mathrm{I}\right)$.

\subsection{Diffusion Models for Video Generation}

Applying diffusion models to video generation requires designing the
network $\epsilon_{\theta}$ to not only capture temporal coherence
but also scale effectively to long sequences. Ho et al. \cite{vdm2022}
address this by employing a factorized 3D U-Net architecture, which
first applies spatial attention within each frame and then temporal
attention across tokens at corresponding spatial locations. Meanwhile,
Ma et al. \cite{ma2025latte} adopt Vision Transformers \cite{dosovitskiy2020image}
as the backbone for $\epsilon_{\theta}$ and systematically evaluate
different spatial-temporal factorization strategies, including variants
proposed in prior work \cite{lin2024open}. Their experiments show
that the spatial-temporal attention similar to the one in \cite{vdm2022}
achieves the best results. However, restricting temporal attention
to tokens at the same spatial locations is less effective when handling
large motions between frames, since moving objects rarely stay aligned
spatially \cite{yang2025cogvideox}. Therefore, subsequent works such
as \cite{yang2025cogvideox,wan2025wan,sun2025ar,song2025history}
leverage Full 3D Attention over all spatio-temporal tokens to achieve
better temporal coherence.

\section{Method}

In this section, we provide a detailed explanation of our proposed
method, as illustrated in \Figref{overview}. The core idea of $\Model$
is Matrix Attention, a frame-level temporal attention mechanism designed
to efficiently model the complex spatio-temporal dependencies inherent
in video data. We first detail the formulation of Matrix Attention
and then describe how to integrate it into DiT.

\subsection{Matrix Attention}

A fundamental challenge in video diffusion is modeling the intricate
relationships between spatial and temporal dimensions. As discussed
in \Secref{Introduction}, existing Factorized approaches typically
restrict temporal attention to tokens at identical spatial locations
across frames. While this design reduces computational cost compared
to Full 3D Attention, it also increases learning complexity and makes
it difficult to ensure object-level consistency across frames. To
overcome these limitations, we introduce \emph{Matrix Attention},
a frame-level attention mechanism that operates at the \emph{frame
level} rather than the token level. The key idea is to represent each
frame $z^{t}$ as a matrix, $z^{t}\in\Real^{N\times D}$ where $N$
is the number of tokens per frame and $D$ is the feature dimensionality
of each token, and leverage matrix-native operations to compute similarities
between frames.

First, we map $z^{t}$ to query, key, and value matrices $q^{t},k^{t}\in\Real^{N_{qk}\times D_{qk}}$,
$v^{t}\in\Real^{N_{v}\times D_{v}}$ using corresponding matrix-native
operations as follows:
\begin{align}
q^{t} & =U_{q}^{\intercal}z^{t}W_{q}+B_{q},\\
k^{t} & =U_{k}^{\intercal}z^{t}W_{k}+B_{k},\\
v^{t} & =U_{v}^{\intercal}\latent^{t}W_{v}+B_{v},\label{eq:matrix-linear}
\end{align}
where $U_{q},U_{k}\in\Real^{N\times N_{qk}}$, $U_{v}\in\Real^{N\times N_{v}}$,
$W_{q},W_{k}\in\Real^{D\times D_{qk}}$, $W_{v}\in\Real^{D\times D_{v}}$
are learnable row- and column-weight matrices; $B_{q},B_{k}\in\Real^{N_{qk}\times D_{qk}}$
and $B_{v}\in\Real^{N_{v}\times D_{v}}$ are bias matrices. It is
worth noting that each row of $q^{t}$, $k^{t}$, and $v^{t}$ contains
combined information between all tokens in the frame $z^{t}$, allowing
to capture different aspects at the frame level. Applying this to
all frames $z:=\left[z^{t}\right]_{t=1}^{T}\in\Real^{T\times N\times D}$
in a video, we obtain $q:=\left[q^{t}\right]_{t=1}^{T}$, $k:=\left[k^{t}\right]_{t=1}^{T}$
$\in\Real^{T\times N_{qk}\times D_{qk}}$, and $v:=\left[v^{t}\right]_{t=1}^{T}$
$\in\Real^{T\times N_{v}\times D_{v}}$. The attention output is defined
as:
\begin{equation}
u=\text{MatrixAttention}\left(\query,\key,\mvalue\right)=\text{Softmax}\left(\text{Sim}\left(\query,\key\right)\right)\cdot\mvalue.
\end{equation}
Here, $S:=\text{Sim}\left(q,k\right)\in\Real^{T\times T}$ is a similarity
matrix where each entry $S^{t,t'}$ measures the similarity between
two matrices $q^{t}$, $k^{t'}$, and can be computed via the scaled
Frobenius inner product

\begin{equation}
S^{t,t'}=\frac{\langle q^{t},k^{t}\rangle_{\text{F}}}{\sqrt{N_{qk}D_{qk}}},
\end{equation}
where $\langle q^{t},k^{t}\rangle_{\text{F}}$ is the Frobenius
inner product between $q^{t}$ and $k^{t}$. To enhance expressiveness,
\emph{Multi-head Matrix Attention} can be obtained by splitting $q$,
$k$, and $v$ along their row and column dimensions, and applying
standard Matrix Attention to each partition $(i,j)$:
\begin{equation}
u_{i,j}=\text{MatrixAttention}\left(\query_{i,j},\key_{i,j},\mvalue_{i,j}\right),
\end{equation}
where $\query_{i,j},\key_{i,j}\in\Real^{T\times\frac{N_{qk}}{m}\times\frac{D_{qk}}{n}}$,
$v_{i,j}\in\Real^{T\times\frac{N_{v}}{m}\times\frac{D_{v}}{n}}$,
and $m$, $n$ denote the number of row and column splits, respectively.
The final output $u$ is reconstructed by concatenating all partitions
$u_{i,j}$.

\subsection{FrameDiT}

Our $\Model$ architecture is designed based on DiT with interleaved
spatial and temporal blocks, as shown in \Figref{overview}. This
decoupling is critical for computational efficiency. While the spatial
blocks are left unchanged, the temporal blocks are designed with our
Matrix Attention. We propose and analyze two variants that represent
a trade-off between simplicity and expressive power.

\paragraph{$\boldsymbol{\protect\Modelglobal}$: Global-only}

We replace the local temporal attention with our proposed Matrix Attention,
enabling the model to perform temporal attention at the frame level.
This configuration serves to isolate the effectiveness of the global,
frame-level context provided by Matrix Attention. 

\paragraph{$\boldsymbol{\protect\Modeluniversal}$: Global-Local Hybrid}

Recognizing that temporal dynamics in video exist at multiple scales
- from fine-grained pixel motion to high-level scene changes, our
$\Model$ architecture employs a Global-Local Hybrid approach. This
variant uses two parallel attention branches, spatially local temporal
attention to capture fine-grained motion and local consistency, and
Matrix Attention operates on the frame matrices, capturing frame-level
information, and object-level consistency across distant spatial positions.
We fuse output of two branches $e_{\text{local}}$, $e_{\text{global}}$
by concatenating and projecting them by a linear layer.
\begin{equation}
e=\text{MLP}\left(\text{concat}\left(e_{\text{local}},e_{\text{global}}\right)\right)
\end{equation}

\paragraph{Complexity}

The computational cost of $\Modelglobal$ is $O\left(TN^{2}+T^{2}N_{qk}\right)$,
coming from spatial attention and our frame-level Matrix Attention,
while $\Modeluniversal$ adds spatially local temporal attention on
top $O\left(TN^{2}+T^{2}N+T^{2}N_{qk}\right)$. When $N_{qk}\ll N$,
the $T^{2}N_{qk}$ term is negligible, making complexity of $\Modeluniversal$
is nearly identical to the Local Factorized Attention. In practice,
for high-resolution videos, spatial attention dominates, and both
variants match the efficiency of factorized designs while adding global
temporal context at minimal overhead. For long sequences, temporal
attention dominate, and both models maintain the same efficiency while
avoiding quadratic cost of full 3D attention.

\begin{figure*}[!tp]
\begin{centering}
\resizebox{0.95\linewidth}{!}{ \includegraphics[width=1\linewidth]{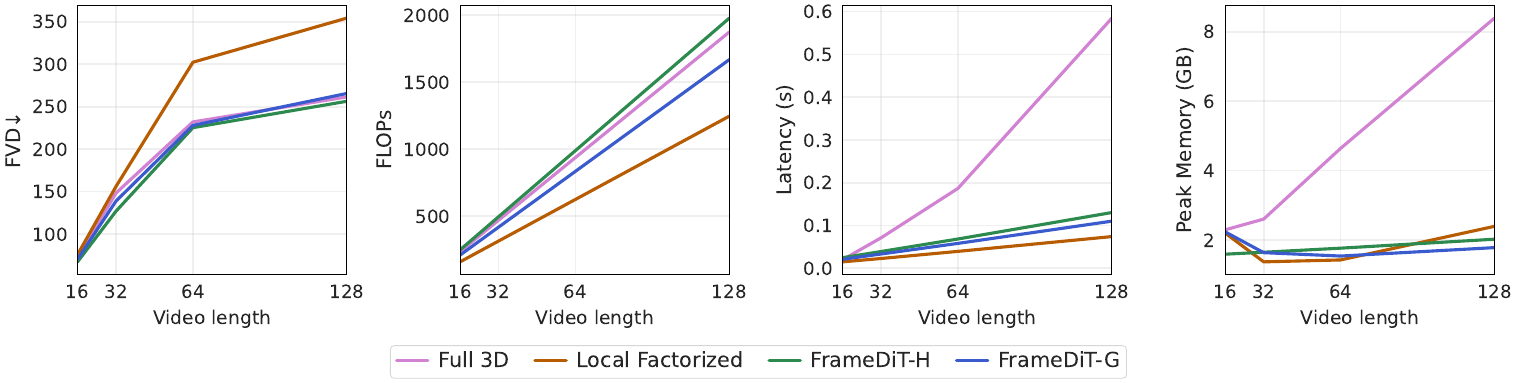}}
\par\end{centering}
\caption{\textbf{Scaling with video length.} We compare Local Factorized, Full
3D attention, and our $\protect\Model$ variants as video length increases
from 16 to 128 frames on the $128\times128$ Taichi dataset. From
left to right: FVD, FLOPs, inference latency, and peak memory. While
Full 3D achieves competitive quality, it exhibits steep growth in
computational and memory costs. In contrast, our models maintain comparable
or better FVD while scaling more efficiently, with latency and memory
close to Local Factorized attention. \label{fig:metrics-vs-frames}}
\end{figure*}

\subsection{Integrating Matrix Attention into existing Video DiTs}

In this section, we describe how Matrix Attention can be integrated
into existing video DiT architectures. We focus on Local Factorized
models, including Latte\cite{ma2025latte}, OpenSora 1.3 \cite{zheng2024open},
and OpenSora-Plan 1.1 \cite{lin2024open}. 

Our primary design introduces Matrix Attention as an additional temporal
branch alongside the original Local Factorized temporal attention.
The outputs of the two branches are fused via a learnable gating mechanism.
This hybrid design preserves the pretrained temporal modeling capacity
while enabling global temporal interaction through Matrix Attention.
We explore both softmax and concatenation gate. For softmax gate,
we initialize the gate such that the pretrained Local Factorized branch
receives a dominant weight $(\approx0.97)$, while Matrix Attention
receives a small initial weight $\left(\approx0.03\right)$. This
ensures the model behavior at early training closely matches the pretrained.
However, both the softmax gate and the Matrix Attention parameters
change minimally during training. We attribute this to small gradient
value induced by softmax: when one branch dominates, the gradient
flowing to the suppressed branch becomes extremely small, making it
difficult for Matrix Attention to learn effectively. Therefore, the
hybrid model fails to exploit the additional temporal modeling capacity.

To mitigate this problem, we adopt the concatenation through a linear
layer. It is initialized using Kaiming initialization \cite{he2015delving}
for weights and zero for bias. Compared to softmax, it avoids early
saturation and enables more balanced gradient flow between branches.
Empirically, it leads to stable training and consistent performance
gains.

We further investigate a more aggressive design in which the original
Local Factorized temporal attention is completely removed and replaced
by Matrix Attention. Although after training, this configuration successfully
follows prompts and produces visually plausible frames, the generated
frames resemble a sequence of independent images rather than a coherent
video. We hypothesize that the pretrained Local Factorized temporal
attention encodes strong motion priors that are difficult to recover
when replaced entirely. Since Matrix Attention is newly initialized
and trained from scratch, removing the pretrained local factorized
temporal attention eliminates critical inductive bias and destabilizes
temporal coherence. This observation motivates our hybrid design,
which preserves existing local branch while improving it with global
temporal interaction.

\textbf{}

\section{Related Work}

Video generation aims to synthesize realistic videos that exhibit
both high-quality appearance and coherent motion. Early approaches
use generative adversarial networks (GANs), such as MoCoGAN \cite{tulyakov2018mocogan},
DIGAN \cite{yu2022generating}, StyleGAN-V \cite{skorokhodov2022stylegan},
and MoStGAN-V \cite{shen2023mostgan}, which model temporal dynamics
through recurrent or convolutional architectures. While these methods
demonstrate promising modeling, GAN-based video generators suffer
from mode collapse, training instability, and difficulty scaling to
higher resolutions or longer durations. Another line of work employs
autoregressive video transformers such as VideoGPT \cite{yan2021videogpt},
MAGVIT \cite{yu2023magvit}, VideoPoet \cite{kondratyuk2023videopoet}
which model videos as sequences of discrete tokens. These approaches
are effective for reconstruction and prediction tasks, but often struggle
to scale to high-resolution, high-fidelity synthesis due to their
sequential generation.

With the success of diffusion models in image generation, recent research
has extended diffusion-based methods to video. Early video diffusion
models \cite{vdm2022,blattmann2023stable} employ 3D U-Nets with factorized
spatio-temporal attention, where spatial and temporal attentions are
applied separately. Later works \cite{blattmann2023stable,blattmann2023align,yu2023video,rombach2022high}
improve scalability by moving from pixel space to latent space, enabling
larger models and longer sequences. Later on, the strong scalability
of Diffusion Transformers (DiTs) \cite{peebles2023scalable} in image
generation has inspired their adoption for video. These models \cite{ma2025latte,zheng2024open,gupta2024photorealistic,lu2023vdt,hong2022cogvideo,singer2022make}
commonly employ Local Factorized Attention, allowing the reuse of
pretrained text-to-image backbones and finetuning on video data. However,
this formulation assumes motions alignment across frames. As a result,
the model must rely on heavy implicit transmission between layers,
and fails to maintain consistency of objects.

To address this limitations, several recent video diffusion models
\cite{wan2025wan,yang2025cogvideox,song2025history,sun2025ar,chen2025skyreels,kong2024hunyuanvideo,hacohen2024ltx,lin2024open}
adopt full 3D attention, which treats video as a sequence of space-time
tokens and applies joint attention. This improves temporal coherence
and motion stability, but comes at a significant computational cost,
scaling quadratically with both temporal length and spatial resolution.
Consequently, Full 3D Attention is difficult to apply to longer videos
or higher resolutions.

A complementary line of work aims to improve efficiency of Full 3D
Attention. One approach \cite{ding2025efficient,zhang2025faster}
is limiting attention to local spatial or temporal neighborhoods,
motivated by the observation that attention maps tend to be sparse,
but they typically depend on pretrained Full 3D Attention models and
sacrifice global context awareness. Other approaches \cite{chen2025sana,ghafoorian2025attention}
explore linear attention to reduce quadratic complexity, enabling
long video generation, but often struggle to match the expressiveness
of standard attention or require complex training pipelines.

These developments highlight an open important challenge: how to achieve
the strength of Full 3D Attention while retaining the efficiency of
Factorized Attention.

\section{Experiment}

\subsection{Experimental Setup\label{subsec:Experimental-Setup}}

\paragraph{Datasets}

Following prior studies \cite{ma2025latte,sun2025ar}, we conduct
experiments on four video datasets: UCF-101 \cite{soomro2012ucf101},
Sky-Timelapse \cite{xiong2018learning}, FaceForensics \cite{rossler2018faceforensics},
and Taichi-HD \cite{rossler2018faceforensics}. During training, we
randomly sample 16-frame clips from each video using frame interval
of 3. For the 128$\times$128 Taichi-HD setting, we train with longer
clips up to 128 frames to further evaluate long-range temporal coherence.

\vspace{-15pt}

\paragraph{Evaluation metrics}

We evaluate video quality and temporal consistency using Fréchet Video
Distance (FVD) \cite{unterthiner2018towards}, Fréchet Video Motion
Distance (FVMD) \cite{liu2024fr}. For frame-level, we report Fréchet
Inception Distance (FID) \cite{fid}. All metrics are computed over
2,048 generated video clips.

\vspace{-15pt}

\paragraph{Training Setting}

Models at $128\times128$ are trained from scratch with global batch
size 16 for 150K steps. With $N=64$, we choose $N_{qk}=32$, $N_{v}=256$
and and use 32 attention heads along the column dimension. Each experiment
requires approximately 54 H100 GPU hours. Models at $256\times256$
are trained from scratch for 200k steps with $N=256,N_{qk}=128,N_{v}=512$,
and 128 column attention heads, requiring about 280 H100 GPU hours
per experiment. Training uses AdamW with learning rate $1e\!-\!4$.
We employ standard training techniques, including exponential moving
average with a decay rate of 0.999, gradient clipping, and noise clipping. 

\subsection{Video Generation}

\subsubsection{Comparison between attention mechanisms}

\Figref{metrics-vs-frames} presents a comparison of FVD, FLOPs, latency,
and peak memory between our proposed FrameDiT-G/H and other DiT variants
employing Local Factorized attention (LF-DiT) and Full 3D attention
(Full3D-DiT) on the 128$\times$128 Taichi-HD dataset, with video
lengths ranging from 16 to 128 frames. FrameDiT-G/H consistently outperforms
LF-DiT and achieves performance comparable to Full3D-DiT across all
video lengths, demonstrating the effectiveness of the proposed Matrix
Attention.

In terms of computational efficiency, Full3D-DiT exhibits a sharp
increase in FLOPs, latency, and memory overhead as video length grows,
quickly becoming prohibitively expensive. In contrast, FrameDiT-G/H
scales much more gracefully, maintaining latency and memory overhead
close to LF-DiT while delivering substantially improved generation
quality. Overall, FrameDiT-G/H achieves a strong balance between effectiveness
and efficiency.

\subsubsection{Comparision with existing video generative models}

We further compare FrameDiT-G/H with existing GAN-based \cite{yu2022generating,skorokhodov2022stylegan,tulyakov2018mocogan}
and diffusion-based \cite{he2022latent,yu2023video,mei2023vidm,liu2024redefining,chen2024diffusion,ma2025latte}
video generation methods for 16-frame video synthesis at 256$\times$256
resolution, following the evaluation protocol of \cite{skorokhodov2022stylegan,he2022latent,ma2025latte}.
The diffusion-based baselines cover diverse architectural designs,
including U-Net--based models (e.g., LVDM \cite{he2022latent}, PVDM
\cite{yu2023video}), DiTs with Local Factorized Attention (e.g.,
FVDM \cite{liu2024redefining}, Latte \cite{ma2025latte}), and DiT
variants employing causal Full 3D Attention (e.g., AR-Diffusion \cite{sun2025ar}). 

During our experiments, we observed that the reported performance
of AR-Diffusion on Taichi-HD appeared \emph{unusually strong} - showing
roughly a 32\% improvement in FVD over the next-best method Latte.
To validate this result, we re-evaluated AR-Diffusion using its publicly
available official checkpoints\footnote{Link: \url{https://github.com/iva-mzsun/AR-Diffusion}}
under our evaluation protocol. Interestingly, the reproduced results
yielded \emph{substantially higher} (i.e., \emph{worse}) FVD scores
than reported in the original paper, specifically 100.9 vs. 66.3 on
Taichi-HD and 84.0 vs. 71.9 on FaceForensics. For fairness, all comparisons
with AR-Diffusion are based on our reproduced results.

\begin{table}
\caption{\textbf{Comparison of our FrameDiT-G/H with baselines for generating
16-frame videos at $\boldsymbol{256\times256}$ resolution}. Baseline
results are reported from their original papers. {*} indicates results
obtained using the official AR-Diffusion checkpoints.\label{tab:fvd_main}}

\centering{}\resizebox{0.99\linewidth}{!}{ %
\begin{tabular}{lcccc}
\toprule 
\multirow{1}{*}{Model} & UCF101 & Sky & \multicolumn{1}{c}{Taichi-HD} & \multicolumn{1}{c}{Face}\tabularnewline
\midrule 
MoCoGAN \cite{tulyakov2018mocogan} & 2886.9 & 206.6 & - & 124.7\tabularnewline
DIGAN \cite{yu2022generating} & 1630.2 & 83.1 & 128.1 & 62.5\tabularnewline
StyleGAN-V \cite{skorokhodov2022stylegan} & 1431.0 & 79.5 & 143.5 & 47.4\tabularnewline
VideoGPT \cite{yan2021videogpt} & 2880.6 & 222.7 & - & 185.9\tabularnewline
PVDM \cite{yu2023video} & 343.6 & 55.4 & 540.2 & 355.9\tabularnewline
LVDM \cite{he2022latent} & 372.9 & 95.2 & 99.0 & -\tabularnewline
VIDM \cite{mei2023vidm} & 294.7 & 57.4 & 121.9 & -\tabularnewline
FVDM \cite{liu2024redefining} & 468.2 & 106.1 & 194.6 & 55.0\tabularnewline
Diffusion Forcing \cite{chen2024diffusion} & 274.5 & 251.9 & 202.0 & 99.5\tabularnewline
Latte \cite{ma2025latte} & 202.2 & 42.7 & 97.1 & 27.1\tabularnewline
AR-Diffusion \cite{sun2025ar} & 186.6 & 40.8 & 66.3 & 71.9\tabularnewline
AR-Diffusion{*}  & \uline{181.9} & \uline{40.2} & 100.9 & 84.0\tabularnewline
\midrule
\textbf{$\Modelglobal$} & 201.6 & 40.6 & \uline{96.8} & \uline{21.5}\tabularnewline
\textbf{$\Modeluniversal$} & \textbf{170.1} & \textbf{39.5} & \textbf{95.5} & \textbf{16.6}\tabularnewline
\bottomrule
\end{tabular}}
\end{table}

As shown in Table~\ref{tab:fvd_main}, FrameDiT-G consistently outperforms
Latte across all datasets while maintaining comparable computational
efficiency, demonstrating the advantage of global Matrix Attention
over Local Factorized Attention for video modeling. Nevertheless,
FrameDiT-G still trails AR-Diffusion on UCF101 and Sky-Timelapse,
indicating that relying solely on frame-level global attention is
insufficient for fully realistic video synthesis. By incorporating
hybrid global-local attention, FrameDiT-H further improves performance,
surpassing all baselines and establishing new state-of-the-art results
across all datasets. In particular, it achieves approximately a 9\%
improvement in FVD over AR-Diffusion on UCF101 and a 39\% gain over
Latte on FaceForensics, highlighting its effectiveness in modeling
complex motion patterns, fast-moving foreground objects, and slowly
evolving backgrounds.

\begin{figure}
\begin{centering}
\includegraphics[width=0.98\columnwidth]{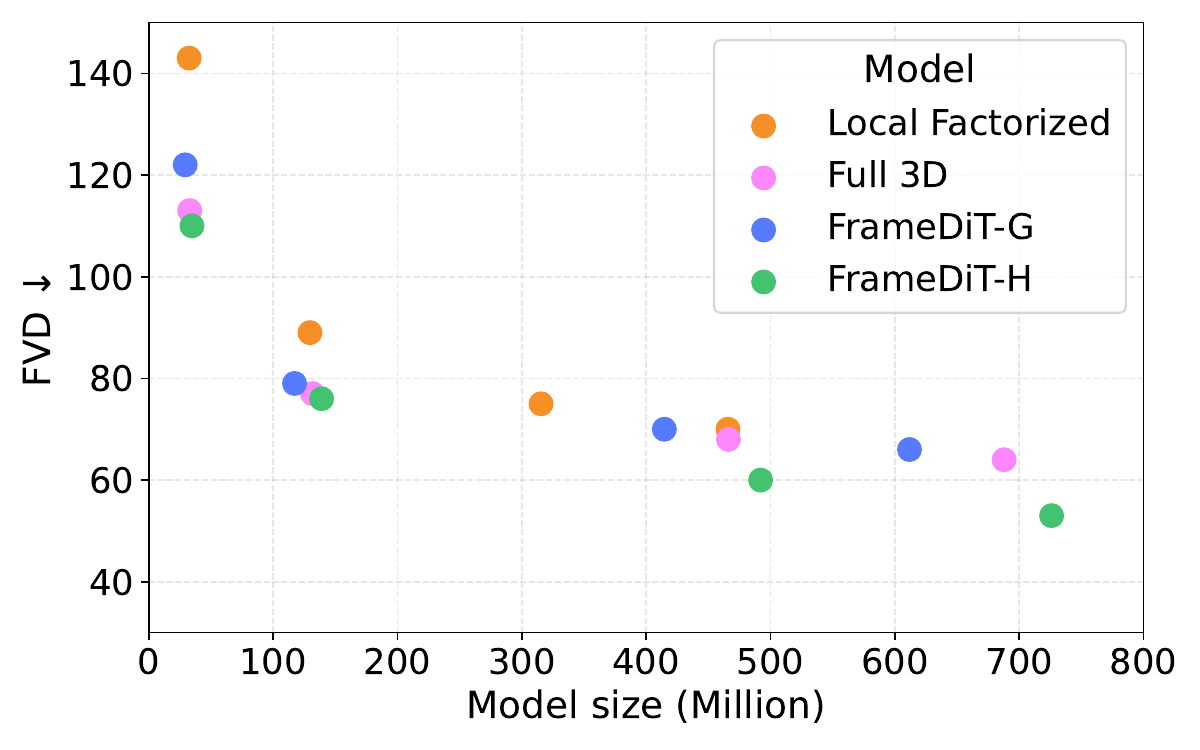}
\par\end{centering}
\caption{\textbf{FVD comparison of different models as increasing model size}.
Each bubble shows a model variant, where y-axis reports FVD, and bubble
diameter is proportional to GFLOPs. \label{fig:model-size}}
\end{figure}

\begin{table*}[tp]
\centering{}\caption{\textbf{Performance comparison of various text-to-video generation
models on the VBench benchmark.}\label{tab:t2v_results}}
\resizebox{0.99\linewidth}{!}{ %
\begin{tabular}{cc>{\centering}p{1.7cm}>{\centering}p{1.4cm}>{\centering}p{1.4cm}>{\centering}p{1.5cm}>{\centering}p{1.9cm}>{\centering}p{1.9cm}>{\centering}p{1.6cm}>{\centering}p{1.9cm}>{\centering}p{1.5cm}>{\centering}p{1.5cm}}
\toprule 
Model & \#Params & Attention Type & Total Score & Quality Score & Semantic Score & Subject Consistency & Background Consistency & Temporal Flickering & Motion Smoothness & Dynamic Degree & Temporal Style\tabularnewline
\midrule
Latte \cite{ma2025latte} & 1B & Factorized & 77.29 & 79.72 & 67.58 & 88.88 & 95.40 & 98.89 & 94.63 & 68.89 & 24.76\tabularnewline
Lavie \cite{wang2025lavie} & 3B & Factorized & 77.08 & 78.78 & 70.31 & 91.41 & 97.47 & 98.30 & 96.38 & 49.72 & 25.93\tabularnewline
OpenSoraPlan 1.1 \cite{lin2024open} & 1B & Factorized & 78.00 & 80.91 & 66.38 & 95.73 & 96.73 & 99.03 & 98.28 & 47.72 & 23.87\tabularnewline
OpenSoraPlan 1.3 \cite{lin2024open} & 2.7B & Full 3D & 77.23 & 80.14 & 65.62 & 97.79 & 97.24 & 99.20 & 99.05 & 30.28 & 22.47\tabularnewline
LTX-Video \cite{hacohen2024ltx} & 1.9B & Full 3D & 80.00 & 82.30 & 70.79 & 96.56 & 97.20 & 99.34 & 98.86 & 54.35 & 22.62\tabularnewline
Wan 2.1 \cite{wan2025wan} & 2B & Full 3D & 84.26 & 85.30 & 80.09 & 96.34 & 97.29 & 99.49 & 97.44 & 85.56 & 25.31\tabularnewline
\midrule 
$\Modeluniversal$ & 1.3B & Matrix & 79.12 & 81.69 & 68.84 & 95.10 & 96.37 & 97.16 & 95.97 & 70.83 & 24.84\tabularnewline
\bottomrule
\end{tabular}}
\end{table*}

\subsection{Text-to-Video Generation}

\begin{table}
\caption{\textbf{Comparison of the normalization for row-weight matrix $U$.}
Result shows that softmax achieves the best overall performance across
FVD, FVMD, and FID metrics.\label{tab:matrix-u-design}}

\centering{}\resizebox{0.75\linewidth}{!}{ %
\begin{tabular}{lccc}
\toprule 
\multirow{1}{*}{$U$ normalization} & FVD$\downarrow$ & FVMD$\downarrow$ & FID$\downarrow$\tabularnewline
\midrule 
No & 70.31 & 990.00 & 14.44\tabularnewline
Softmax & \textbf{66.15} & \textbf{943.32} & \textbf{13.45}\tabularnewline
$\ell_{1}\text{-normalization}$ & 66.79 & 987.43 & 13.83\tabularnewline
$\ell_{2}\text{-normalization}$ & 67.13 & 984.72 & 13.44\tabularnewline
\bottomrule
\end{tabular}}
\end{table}

We evaluate the effectiveness of Matrix Attention on real-world text-to-video
(T2V) generation by building upon a pretrained Latte model for T2V
obtained from \cite{ma2025latte}. Specifically, we augment each Local
Factorized temporal attention layer in Latte with an additional Matrix
Attention module, forming FrameDiT-H. With the original number of
tokens per frame is $N=1024$, we set $N_{qk}=1$, and $N_{v}=2$
and choose 512 column attention heads, resulting the model $\Modeluniversal$
with additional 314M parameters from 1B Latte model. During training,
all pretrained Latte parameters are kept frozen, and only the newly
introduced Matrix Attention modules are updated. The resulting model
is trained on the Pexels-400K dataset \cite{pexels} for 100K steps
with a global batch size of 8 at 512$\times$512 resolution. Total
training time costs 480 hours of H100 gpu. 

We benchmark FrameDiT-H against recent T2V approaches that use either
Local Factorized \cite{lin2024open,ma2025latte,wang2025lavie} or
Full 3D Attention \cite{lin2024open,hacohen2024ltx,wan2025wan} on
the VBench \cite{huang2024vbench} - which provides a comprehensive
evaluation across multiple dimensions, including visual fidelity,
semantic alignment, temporal coherence, and motion dynamics.

As shown in Table~\ref{tab:t2v_results}, FrameDiT-H consistently
outperforms Latte across several key metrics, including Quality Score
(81.69 vs. 79.72), Semantic Score (68.84 vs. 67.58), Subject Consistency
(95.10 vs. 88.88), Motion Smoothness (95.97 vs. 94.63), and Dynamic
Degree (70.83 vs. 68.89). These improvements stem from the introduced
Matrix Attention modules, which enable more effective modeling of
global video dynamics that are not captured by Latte. \Figref{t2v-qualitative}
shows that our model achieves comparable aesthetic quality to Latte
while significantly improving subject, and background consistency,
and motion smoothness. Compared with other Local-Factorized-Attention-based
models such as Lavie and OpenSora-Plan 1.1, FrameDiT-H achieves substantially
higher Dynamic Degree while maintaining competitive performance in
Quality Score, Semantic Score, Subject Consistency, and Motion Smoothness.
This demonstrates the strength of Matrix Attention in modeling large
and complex motion patterns.

Moreover, our model narrows the gap in performance comparing to Full
3D models. For instance, it achieves a Quality Score of 81.69, close
to LTX-Video\textquoteright s 82.30. Wan 2.1 achieves the highest
Total Score (84.26); however, it benefits from larger and carefully
curated training datasets. In contrast, $\Modeluniversal$ is trained
solely on the publicly available Pexels-400K dataset while keeping
the pretrained backbone frozen. Therefore, the observed gains can
be attributed directly to the Matrix Attention mechanism rather than
increased data scale or full-model retraining.

\subsection{Ablation study}

To validate our design and quantify the contribution of each component,
we conduct many ablation experiments, each is performed on 16-frame
$128\times128$ Taichi-HD dataset.

\paragraph{Normalizations of row-weight matrix $\boldsymbol{U}$ \label{par:Design-of-u}}

We study effect of different normalization strategies applied to the
row-weight matrix $U$, which controls how spatial tokens are synthesized
into the frame-level matrix representation. \Tabref{matrix-u-design}
shows the comparison between four configurations: no normalization,
softmax, and $\ell_{1},\ell_{2}\text{-normalization}$. Among these,
softmax normalization yields the best performance achieving an FVD
of 66.15 and an FVMD of 943.32. This hints that normalizing $U$,
particularly via softmax, ensures the synthesized frame representation
within the original embedding manifold, resulting in more stable temporal
attention and stronger overall performance.

\paragraph{Size of row-weight matrix $U$\label{par:Row-mapping-matrix-size}}

We analyze the impact of $N_{qk}$, number of synthesized key/query
tokens in $\Modelglobal$ by varying it from 1 to 64, with $N=64$.
As shown in \Tabref{matrix-u-size}, performance degrades as $N_{qk}$
decreases, with FVD increasing from 66.15 $(N_{qk}=64)$ to 72.16
($N_{qk}=1$), while FVMD rises from 943.32 to 1042.76, indicating
reduced temporal modeling capacity under stronger compression. Even
in the extreme case of compressing all 64 spatial tokens into a single
row matrix, the model remains stable and maintains reasonable performance.
This observation suggests that the row-weight matrix $U$ can perform
lossy data compression, filtering redundant spatial information within
each frame while preserving sufficient information for temporal attention.
In contrast, increasing $N_{qk}$ improves model performance, while
adding only marginal GFLOPs, confirming that the mechanism provides
a flexible and efficient quality-cost trade-off. Meanwhile, since
spatial attention remains unchanged across configurations, the frame-level
metric FID shows only minor variations as $N_{qk}$ changes.

\begin{table}
\caption{\textbf{Effect of the row-weight dimension $\boldsymbol{N}_{\boldsymbol{qk}}$}.
Smaller $N_{qk}$ provides stronger compression and lower GFLOPs,
while larger$N_{qk}$ consistently improves FVD, FVMD. \label{tab:matrix-u-size}}

\centering{}\resizebox{0.75\linewidth}{!}{ %
\begin{tabular}{ccccc}
\toprule 
\multirow{1}{*}{{\small{}$N_{qk}$}} & {\small{}GFLOPs} & {\small{}FVD$\downarrow$} & {\small{}FVMD$\downarrow$} & {\small{}FID$\downarrow$}\tabularnewline
\midrule 
{\small{}1} & {\small{}341.60} & {\small{}72.16} & {\small{}1042.76} & {\small{}14.47}\tabularnewline
{\small{}2} & {\small{}342.03} & {\small{}71.71} & {\small{}1044.55} & {\small{}14.81}\tabularnewline
{\small{}4} & {\small{}342.89} & {\small{}70.50} & {\small{}1035.34} & {\small{}14.74}\tabularnewline
{\small{}8} & {\small{}344.62} & {\small{}69.41} & {\small{}1000.23} & {\small{}14.45}\tabularnewline
{\small{}16} & {\small{}348.07} & {\small{}68.21} & {\small{}990.02} & {\small{}14.30}\tabularnewline
{\small{}32} & {\small{}354.96} & {\small{}67.40} & {\small{}959.91} & {\small{}14.01}\tabularnewline
{\small{}64} & {\small{}368.75} & {\small{}66.15} & {\small{}943.32} & {\small{}13.45}\tabularnewline
\bottomrule
\end{tabular}}
\end{table}

\paragraph{Scaling model size}

We train Local Factorized, Full Attention, and our FrameDiT models
over 4 model size configs (S, B, L, XL) on Taichi dataset at $128\times128$
resolution with 16 frames. \Figref{model-size} demonstrates how their
FVD at 150k training steps scales with model capacity. As model size
increases, all methods benefit from additional capacity; however,
our $\Modelglobal$ and $\Modeluniversal$ consistently outperform
the Local Factorized baseline and achieve performance comparable to,
or better than, Full 3D Attention models. Notably, the performance
gap widens at larger scales: our models exhibit larger reductions
in FVD, attaining the best overall result at the XL configuration
while maintaining competitive computational cost.

\section{Conclusion}

In this paper, we introduced Matrix Attention, a frame-level attention
that leverages matrix-native operations to construct queries, keys,
and values, help preserve the inherent spatio-temporal structure of
video data. Building on this mechanism, we proposed $\Modelglobal$
and $\Modeluniversal$, factorized DiT architectures that balance
expressiveness and computational efficiency for video diffusion models.
Extensive experiments across diverse benchmarks demonstrate that $\Modeluniversal$
delivers strong video quality, and competitive or SOTA performance
while maintaining practical efficiency. In future work, we plan to
further explore the design and parameterization of the row-weight
matrix $U$ to enhance temporal representation.

\pagebreak{}

\bibliographystyle{ieeenat_fullname}
\bibliography{reference}

\newpage{}

\appendix

\section{Theoretical Proof of Matrix Attention}

\begin{table*}[t]
\begin{centering}
\resizebox{0.999\textwidth}{!}{ %
\begin{tabular}{ccccccccccccc}
\toprule 
 & \multicolumn{4}{c}{{\footnotesize{}FVD$\downarrow$}} & \multicolumn{4}{c}{{\footnotesize{}FVMD$\downarrow$}} & \multicolumn{4}{c}{{\footnotesize{}FID$\downarrow$}}\tabularnewline
\cmidrule{2-13} \cmidrule{3-13} \cmidrule{4-13} \cmidrule{5-13} \cmidrule{6-13} \cmidrule{7-13} \cmidrule{8-13} \cmidrule{9-13} \cmidrule{10-13} \cmidrule{11-13} \cmidrule{12-13} \cmidrule{13-13} 
 & {\footnotesize{}16} & {\footnotesize{}32} & {\footnotesize{}64} & {\footnotesize{}128} & {\footnotesize{}16} & {\footnotesize{}32} & {\footnotesize{}64} & {\footnotesize{}128} & {\footnotesize{}16} & {\footnotesize{}32} & {\footnotesize{}64} & {\footnotesize{}128}\tabularnewline
\midrule 
{\footnotesize{}$\Modeluniversal$-Concat} & \textbf{\footnotesize{}66.15} & \textbf{\footnotesize{}126.90} & \textbf{\footnotesize{}225.39} & \textbf{\footnotesize{}256.40} & \textbf{\footnotesize{}943.32} & \textbf{\footnotesize{}478.14} & \textbf{\footnotesize{}125.81} & \textbf{\footnotesize{}70.10} & {\footnotesize{}13.45} & {\footnotesize{}15.48} & {\footnotesize{}17.54} & {\footnotesize{}22.29}\tabularnewline
\midrule 
{\footnotesize{}$\Modeluniversal$-Gated} & {\footnotesize{}67.55} & {\footnotesize{}130.88} & {\footnotesize{}233.27} & {\footnotesize{}265.25} & {\footnotesize{}964.13} & {\footnotesize{}528.42} & {\footnotesize{}153.65} & {\footnotesize{}89.23} & \textbf{\footnotesize{}13.14} & \textbf{\footnotesize{}15.29} & \textbf{\footnotesize{}17.50} & \textbf{\footnotesize{}21.42}\tabularnewline
\bottomrule
\end{tabular}}
\par\end{centering}
\caption{\textbf{Comparison between different Fusion layers for $\protect\Modeluniversal$}.
$\protect\Modeluniversal$-Concat is the model using Concat+MLP fusion,
and $\protect\Modeluniversal$-Gated is the model using sigmoid fusion
described in \parref{Fusion-mechanism}. \label{tab:Comparison-fusion}}
\end{table*}

This section derives the attention maps of our Matrix Attention and
compare it with Full 3D and Local Factorized, showing that Local Factorized
is our special case. Let $z=\left[z_{t}\right]_{t=1}^{T}$ denote
video features, where the feature of frame $t$ is $z_{t}\in\Real^{N\times D}$.
We represent index of token as a pair $\left(t,n\right)$ where $t$
denotes the frame index, $n$ denotes the spatial token index within
the frame. After flattening all tokens in temporal-spatial order,
the full attention matrix is $A\in\Real^{\left(TN\right)\times\left(TN\right)}$,
with each element $A\left[\left(t,n\right),\left(t',n'\right)\right]$
specifies how much token $\left(t',n'\right)$ attends to token $\left(t,n\right)$.
We omit softmax, scaling terms, and MLP biases since they do not affect
structure of the attention map. 

Full 3D Attention computes output as
\begin{equation}
y=A_{\text{full}}zW_{v},
\end{equation}
where $y$ is the output, attention map $A_{\text{full}}$ is unconstrained.
As a result, each token $\left(t,n\right)$ can directly attend to
any token $\left(t^{\prime},n^{\prime}\right)$. 

Local Factorized Attention replaces this with spatial and temporal
attention. For each frame $t$, the spatial attention operates only
within that frame

\begin{equation}
x_{t,n}=\sum_{n'}S_{t}\left[n,n'\right]z_{t,n^{\prime}}W_{v}',
\end{equation}
where $x_{t,n}$ is the output of spatial attention at position $\left(t,n\right)$,
$S_{t}\in\Real^{N\times N}$ is the spatial attention map. Stacking
all frames:
\begin{equation}
x=SzW_{v}',\label{eq:fact_attn_spatial}
\end{equation}
witth $S=\text{diag}\left(S_{1},\dots,S_{T}\right)\in\Real^{\left(TN\right)\times\left(TN\right)}$
is the block diagonal matrix, meaning no information exchange across
frames. Then, temporal attention compute output for each spatial index
$n$ as
\begin{equation}
y_{t,n}=\sum_{t'}H_{n}\left[t,t'\right]x_{t^{\prime},n}W_{v}'',
\end{equation}
$H_{n}\in\Real^{T\times T}$ is the temporal attention map. We can
also rewrite for all spatial indexes in the matrix form:
\begin{align}
y & =xW_{q}W_{k}^{T}x^{T}xW_{s}''=HxW_{v}'',\label{eq:spatial_attn_temporal}
\end{align}
where $H\in\Real^{\left(TN\right)\times\left(TN\right)}$ is attention
matrix, defined by 
\begin{equation}
\begin{cases}
H\left[\left(t,n\right),\left(t^{\prime},n^{\prime}\right)\right]=H_{n}\left[t,t^{\prime}\right] & \text{if }n=n^{\prime}\\
H\left[\left(t,n\right),\left(t^{\prime},n^{\prime}\right)\right]=0 & \text{otherwise}.
\end{cases}\text{}\label{eq:h_attn_matrix}
\end{equation}
Replace $x$ in \Eqref{spatial_attn_temporal} by \Eqref{fact_attn_spatial}:
\begin{equation}
y=HSzW_{v}'W_{v}''=A_{\text{fact}}zW_{v},
\end{equation}
which reveals that Local Factorized Attention implicitly assumes attention
matrix must be factorized into $A_{\text{fact}}=HS$. Each element
of $A_{\text{fact}}$ is computed as
\begin{align}
 & A_{\text{fact}}\left[\left(t,n\right),\left(t',n'\right)\right]\\
= & \sum_{i,j}H\left[\left(t,n\right),\left(i,j\right)\right]S\left[\left(i,j\right),\left(t',n'\right)\right]\\
= & H\left[\left(t,n\right),\left(t',n\right)\right]S\left[\left(t',n\right),\left(t',n'\right)\right],
\end{align}
since $H\left[\left(t,n\right),\left(i,j\right)\right]=0$ if $j\neq n$,
and $S\left[\left(i,j\right),\left(t',n'\right)\right]=0$ if $i\neq n^{\prime}$.
Thus, the interaction between tokens $\left(t,n\right)$ and $\left(t',n'\right)$
relies solely on the single intermediate token $\left(t',n\right)$.
\[
\left(t^{\prime},n^{\prime}\right)\xrightarrow[\text{spatial}]{S}\left(t^{\prime},n\right)\xrightarrow[\text{temporal}]{H}\left(t,n\right).
\]
In the $\Modelglobal$ model, spatial attention remains the same $x=SzW_{v}'$,
while temporal attention is replaced by Matrix Attention, which is
computed as 
\begin{align}
y & =\underbrace{U_{q}^{\intercal}xW_{q}}_{\text{key}}\underbrace{W_{k}^{T}xU_{k}}_{\text{query}}\underbrace{U_{v,}^{\intercal}xW_{v}''}_{\text{value}}\\
 & =U_{q}^{\intercal}HU_{k}U_{v}^{\intercal}SzW_{v}'W_{v}''.
\end{align}
Letting $H^{\prime}=U_{q}^{\intercal}HU_{k}U_{v}^{\intercal}$, this
simplifies to
\begin{equation}
y=H^{\prime}SzW_{v}=A_{\text{mat}}zW_{v}.
\end{equation}
One noteworthy point is that $H^{'}$ is not constrained like in \Eqref{h_attn_matrix}.
Each element of $A_{\text{mat}}$ is computed as
\begin{align}
 & A_{\text{mat}}\left[\left(t,n\right),\left(t',n'\right)\right]\\
= & \sum_{i,j}H'\left[\left(t,n\right),\left(i,j\right)\right]S\left[\left(i,j\right),\left(t',n'\right)\right]\\
= & \sum_{j=1}^{N}H'\left[\left(t,n\right),\left(t',j\right)\right]S\left[\left(t',j\right),\left(t',n'\right)\right].
\end{align}
Unlike Local Factorized, the interaction between $\left(t,n\right)$
and $\left(t',n'\right)$ may pass through all spatial tokens $\left\{ t',j\right\} _{j=1}^{N}$.
\[
\left(t^{\prime},n^{\prime}\right)\xrightarrow[\text{spatial}]{S}\left\{ \left(t^{\prime},j\right)\right\} _{j=1}^{n}\xrightarrow[\text{temporal}]{H}\left(t,n\right)
\]
This removes the single-token bottleneck imposed by Local Factorized
Attention while remaining significantly more efficient than Full 3D
Attention. In case when $U_{q}=U_{k}=U_{v}=I$, $H'$ reduces to $H$,
and Matrix Attention collapses exactly to the temporal attention used
in Local Factorized Attention. Thus, Local Factorized Attention naturally
emerges as a special case of our formulation, corresponding to the
identity row-weight matrices of Matrix Attention.

\section{Additional Experiment}

\subsection{Additional Implementation Details}

We train all models using the DDPM framework with 1000 training diffusion
steps. According to \cite{nichol2021improved}, we choose to learn
both mean and variance of the reverse process to improve log-likelihood,
increasing sample quality using the loss function
\begin{equation}
L=L_{\text{simple}}+\lambda L_{\text{vlb}},
\end{equation}
with $\lambda=10^{-3}$.

We employ the noise-parameterization for all models. For the latent
representation, we use the Stable Diffusion 2.0 autoencoder \cite{rombach2022high},
encoding each frame independently with a compression ratio of $\left\{ 1,8\right\} $.
We use this VAE to ensure that improvement is solely from the diffusion
modeling rather than the reconstruction quality of the autoencoder.

For dataset, we preprocess them by center crop, and resize to desire
resolution. We randomly sample with fixed frame interval 3 from videos,
except interval 1 for Taichi-HD $128\times128$ at 128 frames. Our
model and all baselines are implemented using the Latte codebase and
trained under the exact same settings to ensure fair comparison. All
conditioning inputs, including noise levels and class labels, are
injected into the network through AdaLN-Zero layers. For all experiments,
training is performed in FP16 precision for computational efficiency.
We clip maximum norm of gradients to 1.0 from 100,000 training steps
to stabilize training, and exponential moving average (EMA) of the
model weights with a 0.999 decay. To improve high-resolution quality
at $256\times256$, we jointly train on both videos and images, with
8 images for each video. We use the AdamW optimizer with a learning
rate of $1\times10^{-4}$. We use DDPM sampler with 250 sampling steps
for all models. For UCF101, we apply classifier-free guidance with
CFG = 7.0.

We evaluate models using both video-level and frame-level metrics.
For video-level evaluation, we adopt FVD \cite{unterthiner2018towards}
and FVMD \cite{liu2024fr}. FVD computes the Fréchet distance between
feature distributions of real and generated videos, where features
are extracted using a pretrained I3D network \cite{carreira2017quo};
it reflects both overall video quality and temporal coherence. FVMD
focuses specifically on motion consistency: it tracks keypoints using
a pretrained PIPs++ model \cite{zheng2023pointodyssey} to obtain
motion trajectories, then measures the Fréchet distance between the
real and generated motion features. Since FVMD processes videos in
16-frame chunks, longer sequences lead to more chunks, therefore,
produce lower FVMD values. For this reason, we report the relative
FVMD improvement over Local Factorized Attention to better reflect
motion consistency gains across different sequence lengths. For frame-wise
metrics, we report FID \cite{fid}, evaluating the similarity between
real and generated frame distributions using Inception features, providing
a measure of overall image realism.

\subsection{Additional Result}

\paragraph{Comparison between different attention mechanisms}

\begin{figure*}
\begin{centering}
\includegraphics[width=0.99\textwidth]{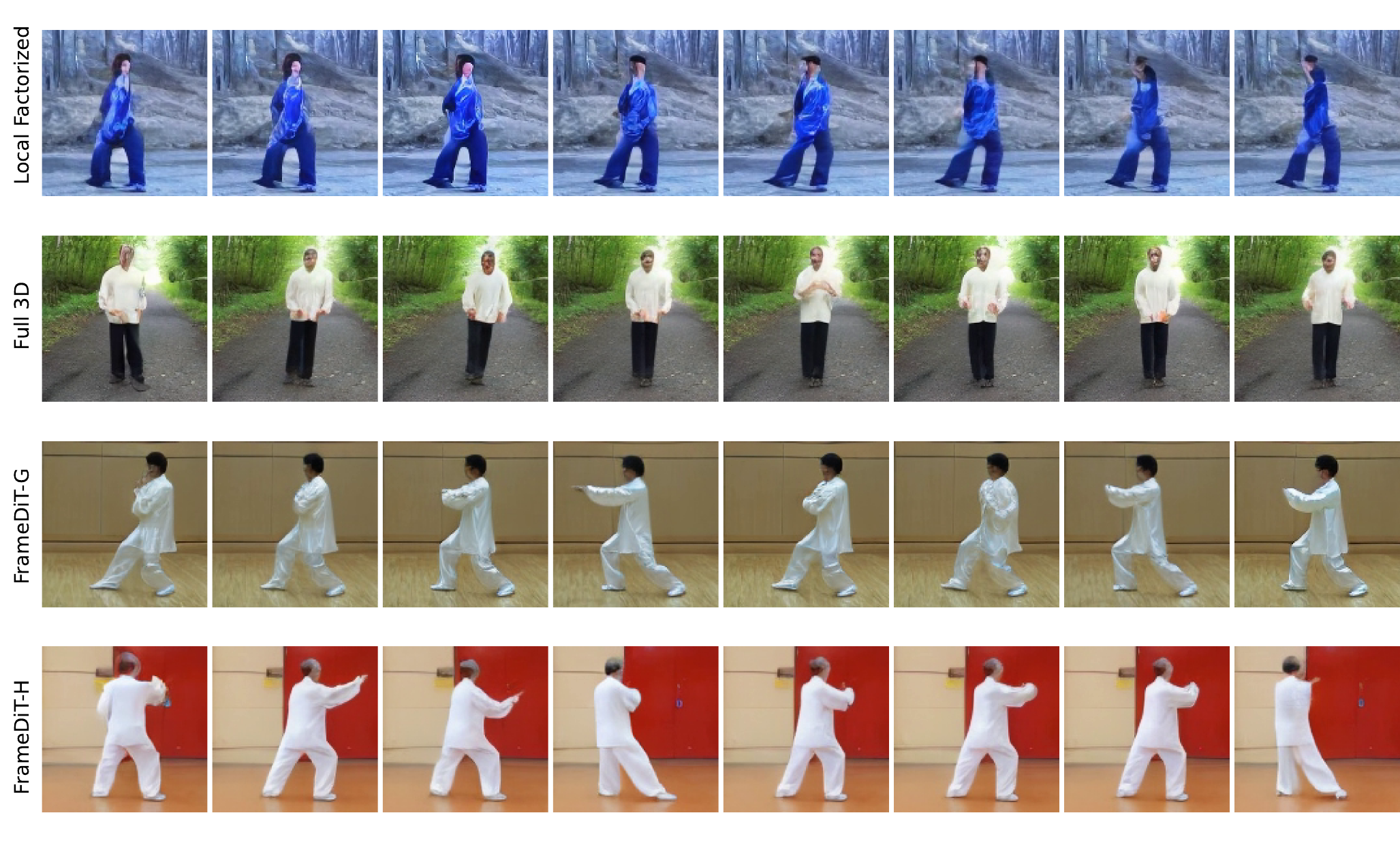}
\par\end{centering}
\caption{\textbf{Qualitative comparison on 128-frame Taichi-HD $\boldsymbol{128\times128}$.
}Local Factorized Attention exhibits severe temporal drift and collapsing
human structure. In contrast, Full 3D model and $\protect\Modelglobal$,
$\protect\Modeluniversal$ remain stable even at 128 frames, generating
smooth and coherent motion. The slight blurring of small regions (hands,
face) arises from the low-resolution encoding of the Stable Diffusion
2.0 autoencoder.\textbf{\label{fig:qualitative-taichi128-128f}}}
\end{figure*}

\Figref{qualitative-taichi128-128f} present qualitative results of
different attention mechanism on Taichi-HD at $128\times128$ resolution
for video lengths of 128 frames. Local Factorized Attention fails
to preserve human structure, producing inconsistent body movements;
these issues worsen as sequence length increases. In contrast, both
Full 3D Attention, $\Modelglobal$ and $\Modeluniversal$ maintain
stable temporal dynamics across all lengths, producing smooth, consistent
human motion even for 128-frame videos.

Because the Stable Diffusion 2.0 autoencoder is trained primarily
on high-resolution datasets, encoding images at $128\times128$ may
degrade small or fine-grained details such as facial features and
hands. As a result, the generated videos may exhibit blurred or distorted
small structures, reflecting a limitation of the underlying Image
VAE.

\paragraph{Comparision with existing video generative models}

\begin{figure*}
\begin{centering}
\includegraphics[width=0.99\textwidth]{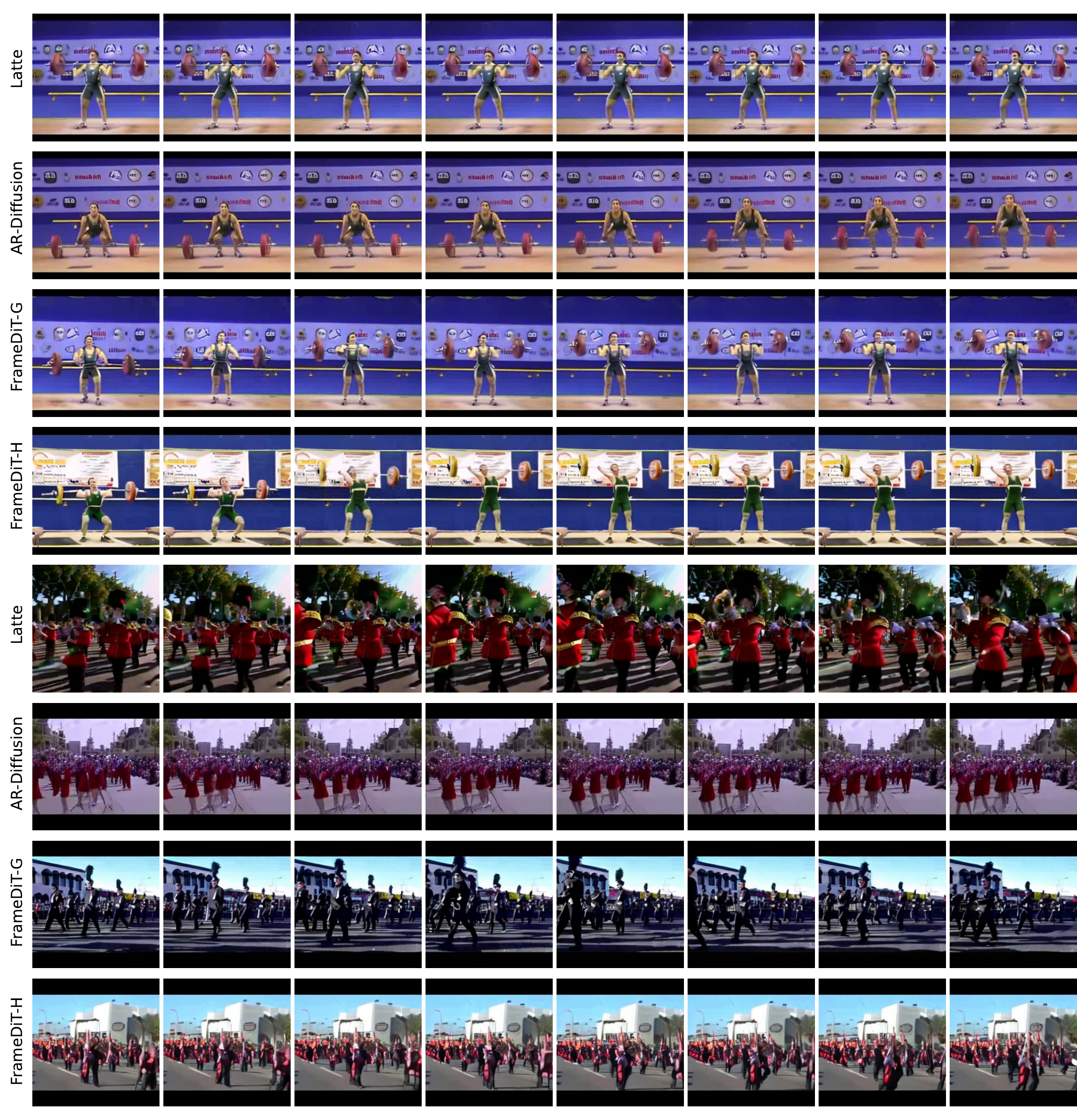}
\par\end{centering}
\caption{\textbf{Qualitative comparison on UCF101 between prior video generative
models and our approach. \label{fig:qualitative-UCF101}}}
\end{figure*}

\Figref{qualitative-UCF101} presents qualitative comparisons on UCF101
dataset between prior video generative models and our approach. Latte
often produces temporally inconsistent videos; for example, in the
second sample of Latte, the scene begins with a single person but
unexpectedly introduces an additional person mid-sequence, revealing
weak long-range temporal modeling. This issue comes from its Local
Factorized Attention, which cannot preserve frame-level information
during temporal attention.

AR-Diffusion produces sharp frames but exhibits minimal motion, partly
due to its use of a frame interval of 1 and independent noise levels
during training. In contrast, our method produces videos with stable
temporal dynamics and maintains coherence even for fast-moving foreground
objects, demonstrating stronger modeling of complex motion and global
temporal structure.

\subsection{Aditional ablation study}

\paragraph{Fusion mechanism \label{par:Fusion-mechanism}}

For $\Modeluniversal$, we investigate alternative fusion mechanisms
that combine local and global temporal features. In addition to the
default Concat+MLP fusion, we evaluate a gated fusion variant where
a learnable scalar gate adaptively controls the contribution of each
branch:

\begin{equation}
e=\sigma\left(\alpha\right)\times e_{\text{local}}+\left(1-\sigma\left(\alpha\right)\right)\times e_{\text{global}},
\end{equation}
where $\sigma\left(\cdot\right)$ denotes the sigmoid function, and
$\alpha$ are learnable parameters. We denote the two variants as
$\Modeluniversal$-Concat and $\Modeluniversal$-Gated, respectively.
As shown in \Tabref{Comparison-fusion}, $\Modeluniversal$-Concat
achieves stronger results on video-level metrics across all sequence
lengths, indicating that preserving the full information from both
branches leads to better temporal modeling. In contrast, $\Modeluniversal$-Gated
performs slightly better on FID, suggesting that gated fusion can
improve frame-wise spatial quality but at the cost of discarding some
temporal information. Overall, the results highlight that both local
and global temporal information are essential, and that aggressively
filtering one branch via a sigmoid gate may hurt video consistency. 

\end{document}